\documentclass[letterpaper, 10pt, conference]{ieeeconf}      

\IEEEoverridecommandlockouts                              

\usepackage{amsmath,amssymb,amsfonts}
\usepackage{algorithm}
\usepackage{algpseudocode}
\usepackage{graphicx}
\usepackage{tabularx}
\usepackage{textcomp}
\usepackage{booktabs}
\usepackage{xcolor}
\usepackage[noadjust]{cite}
\usepackage{tablefootnote}
\usepackage{multirow}
\usepackage{tikz}
\usetikzlibrary{tikzmark}

\definecolor{grn}{rgb}{0.0, 0.5019607843137255, 0.0}
\definecolor{blu}{rgb}{0.0, 0.0, 1.0)}
\definecolor{prp}{rgb}{0.5019607843137255, 0.0, 0.5019607843137255}
\definecolor{dbl}{rgb}{0.0, 0.7490196078431373, 1.0}
\definecolor{white}{rgb}{1.0, 1.0, 1.0}

\def\tablefootnotemark#1{\textsuperscript{\textnormal{\ref{#1}}}}

\usepackage[hidelinks]{hyperref}
\DeclareMathOperator*{\argmin}{argmin}
\title{\LARGE \bf
Data-Efficient Policy Selection for Navigation in Partial Maps via Subgoal-Based Abstraction
}

\author{Abhishek Paudel and Gregory J. Stein
\thanks{Abhishek Paudel and Gregory J. Stein are with Department of Computer Science at George Mason University, Fairfax, Virginia, USA.
({\tt\small \{apaudel4, gjstein\}@gmu.edu})
}
}

\begin{document}
\maketitle


\begin{abstract}
We present a novel approach for fast and reliable policy selection for navigation in partial maps.
Leveraging the recent learning-augmented model-based Learning over Subgoals Planning (LSP) abstraction to plan, our robot reuses data collected during navigation to evaluate how well other \emph{alternative} policies could have performed via a procedure we call \emph{offline alt-policy replay}.
Costs from offline alt-policy replay constrain policy selection among the LSP-based policies during deployment, allowing for improvements in convergence speed, cumulative regret and average navigation cost.
With only limited prior knowledge about the nature of unseen  environments, we achieve at least 67\% and as much as 96\% improvements on cumulative regret over the baseline bandit approach in our experiments in simulated maze and office-like environments.
\end{abstract}

\section{Introduction}

We enable a robot to quickly identify the best performing policy from a set of policies when it is deployed in partially-mapped environments to navigate to an unseen goal.
Goal-directed navigation in partially-mapped environments is an important capability for autonomous robots. A robot tasked with navigating to a faraway goal in an unknown office building will need to build the map as it travels, plan through unknown regions of the map, and decide how to act once that space is revealed.
Efficient navigation in such scenarios requires that the robot be able to reason about the impacts of its actions far into the future, commonly known as the \emph{long-horizon planning} problem, which can be formulated as a Partially-Observable Markov Decision Process (POMDP)~\cite{kaelbling1998planning}.
Solving such POMDPs is computationally intractable, and therefore many existing approaches for planning in such scenarios leverage learning to guide the robot's behavior~\cite{stein2018learning, richter2014high, wayne2018unsupervised_merlin, zhu2017target, kahn2018self, mirowski2016learning}.

When navigating through a partial map, learning often relies on sensor observations collected onboard the robot (e.g., images and laser scans) to make predictions about the goodness of the robot's actions, thus guiding its behavior towards actions that would have resulted in effective performance in its training environments.
However, in general, the robot may have access to a \emph{family} of such learning-informed policies, each trained in a different environment, and must select one from these to guide its behavior during deployment.
After planning with one of these policies, the robot must use this policy's cumulative performance to determine whether it should select that policy again, or switch to another in hopes of reducing navigation cost: an instance of \emph{model selection} applied to \emph{policy selection}.

\begin{figure}
    \centering
    \includegraphics[width=8.0cm]{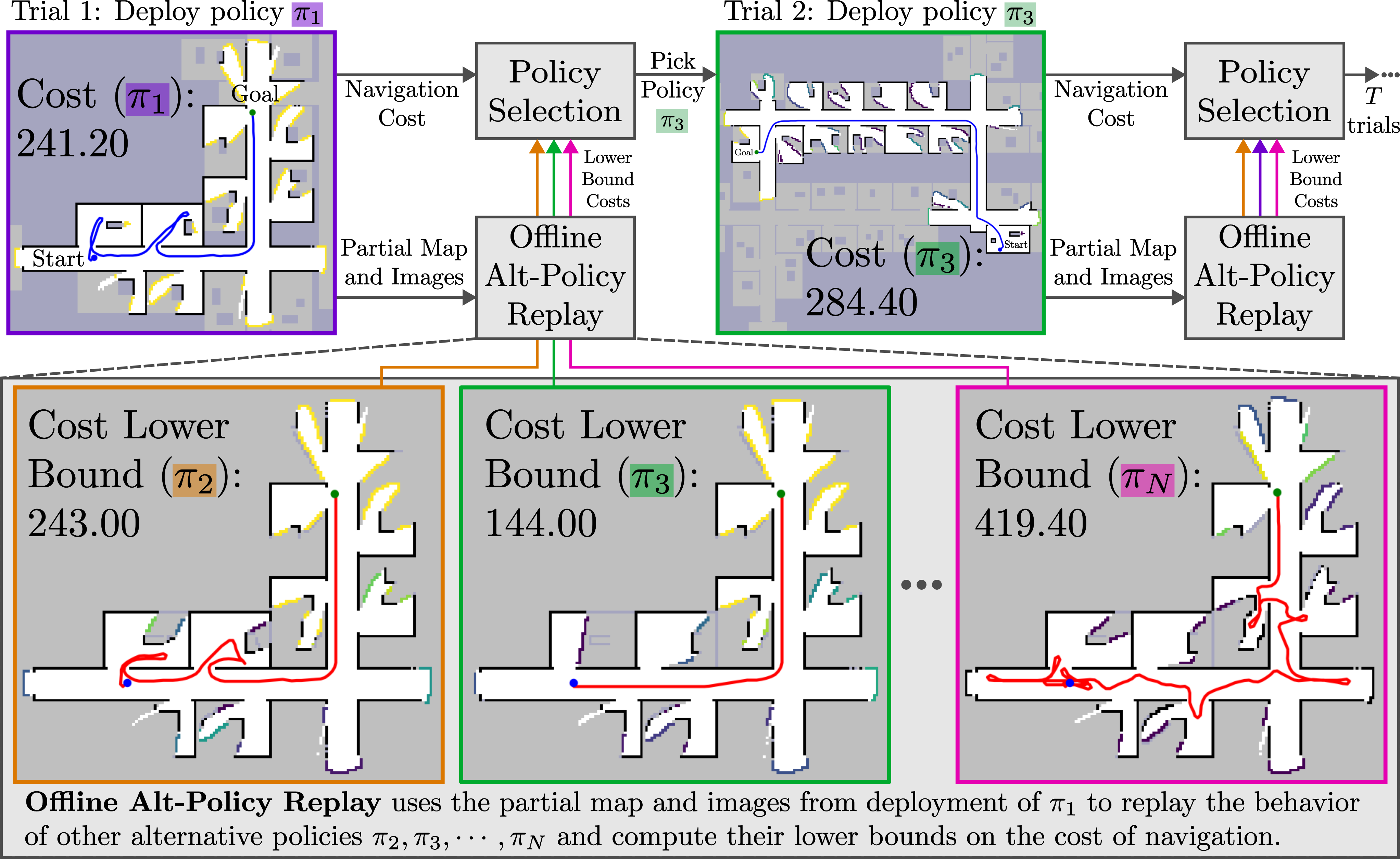}
    \vspace{-0.8em}
    \caption{\textbf{Overview of our approach for data-efficient policy selection for navigation in partial maps.} Our approach relies upon \emph{offline alt-policy replay}, a procedure to compute lower bounds of navigation costs for alternate policies after deployment, bounds used to constrain selection.}
    \vspace{-2.2em}
    \label{fig:overview}
\end{figure}

Model selection approaches~\cite{lee2021online, reisinger2008online, lattimore2020bandit, ucb_lai1985asymptotically, gittins1979bandit, thompson1933likelihood, kuleshov2014algorithms, mannor2004sample} aim to identify the best performing models from a set of pre-defined behaviors by deploying each policy multiple times and evaluating their performance.
Bandit algorithms~\cite{ucb_lai1985asymptotically, thompson1933likelihood, gittins1979bandit}, for example, trade off between \emph{exploitation} (choosing the learning-informed policy that has performed best so far) and \emph{exploration} to pick another policy that \emph{could} improve performance.
Upper Confidence Bound (UCB) Bandits~\cite{ucb_lai1985asymptotically} compute statistical bounds on potential performance for each available policy based on the observed performance of each and how often they have been deployed.
However, bandit algorithms are often black-box, and so tightening these bounds often requires the robot to go through multiple planning attempts and incur poor behavior before such policies can be ruled out.
Owing to the expense associated with lengthy (and poor-performing) navigation trials, data efficiency is critical for such selection approaches to be practically useful.

Instead, we reuse data collected during a trial to evaluate how well other alternative policies could have performed and use this evaluation to tighten the bounds on policy selection: a procedure we call \emph{offline alt-policy replay}, illustrated in Fig.~\ref{fig:overview}.
Based on the information collected during each of its trials, the robot, without collecting additional information from the environment, replays how each of its other alternative policies \emph{could} have performed.
Our offline alt-policy replay approach makes an optimistic assumption about space not seen during the navigation trial (i.e., that the robot will spend only a minimal amount of time in unseen space) allowing us to compute a strict lower bound on the possible performance of each policy.
This information constrains bandit selection---exploration should never select a policy that could not have improved performance during deployment so far---and improves data efficiency of selection.

Our offline alt-policy replay approach depends upon having a planning strategy for which an accurate bound on its would-be performance can be determined via only the limited information collected during the robot’s trials so far.
Not all planning strategies are amenable to such replay, including model-free approaches trained via deep reinforcement learning~\cite{mallozzi2019runtime, kulhanek2019vision}, which can be brittle to changes in their inputs~\cite{henderson2018rlmatters, huang2017adversarial, xiang2018pca}.
It is a key insight of this work that the recent model-based, learning-augmented \emph{Learning over Subgoals Planning} (LSP) of Stein et al.~\cite{stein2018learning} is well-suited for this purpose.
Under the LSP abstraction, learning is used only to make robot-pose-agnostic predictions about statistics of unseen space: e.g., the likelihood that a particular region of space will lead to the unseen goal.
Since its approach to learning is robust to changes in the robot's location, offline replay of LSP policies yields accurate lower bounds on cost that can be used to constrain bandit-like selection.

In this paper, we develop a procedure for data-efficient policy selection called \emph{offline alt-policy replay}, which leverages the information collected during navigation to evaluate how well other alternative policies could have performed and thereby constrain bandit-like selection of a set of Learning over Subgoals Planners~\cite{stein2018learning}.
Our robot is deployed in simulated maze and office-like environments with multiple learning-informed policies: each relies on the Learning over Subgoals Planning abstraction and is trained in a different environment.
We demonstrate that our approach is able to quickly reduce the average navigation cost within much fewer trials compared to baseline UCB bandit approach which generally takes longer to converge and has higher cumulative regret than our approach.
Our results validate that our approach reliably performs better than the UCB bandit and is often able to perform selection much more quickly.

\section{Related Work} \label{sec:related_work}

\textbf{Planning under Uncertainty}\quad{}
POMDPs~\cite{kaelbling1998planning, littman1997} provide is a general framework to represent navigation and exploration under uncertainty~\cite{candido2011minimum, kurniawati2011motion}.
However, they are computationally intractable to solve, and so learning is used for planning~\cite{stein2018learning, richter2014high, wayne2018unsupervised_merlin, zhu2017target, kahn2018self, mirowski2016learning}.
Deep reinforcement learning approaches are also widely used for planning in partially-mapped environments~\cite{gupta2017cognitive, zhang2017deep, tai2017virtualtoreal}, but they are often limited to short-horizon planning.

\textbf{Model Selection}\quad{}
Online model selection approaches in reinforcement learning~\cite{lee2021online, reisinger2008online, ghosh2022model, pacchiano2020regret} generalize bandits and aim to identify best models from a set of available models.
Bandit algorithms~\cite{ucb_lai1985asymptotically, thompson1933likelihood, gittins1979bandit}, which are often black-box, trade off between exploitation and exploration requiring the robot to go through multiple trials with poor behavior before identifying a better policy.

\textbf{Runtime Monitoring}\quad{}
So-called \emph{runtime monitoring} approaches aim to evaluate the deployment-time reliability of learning-guided robot behavior~\cite{Rahman_2021, mallozzi2019runtime, zhou2019automated, daftry2016introspective} to decide whether one should continue using learning or resort to backup strategies in case of degraded performance.
Most such approaches, however, focus on tackling issues associated with shortcomings of local perception or aim to monitor performance in terms of the underlying learned model's raw predictions as opposed to the overall task performance~\cite{zhou2019automated, daftry2016introspective}, thus limiting their effectiveness in evaluating the goodness of long-horizon behavior.

\section{Problem Formulation}

\subsection{Goal-directed Navigation in Partial Maps}
Our robot is placed in a partially-mapped environment and tasked to reach a point-goal in unseen space in minimum expected distance.
We formulate this problem as a Partially Observable Markov Decision Process (POMDP)~\cite{kaelbling1998planning}.
The belief state $b_t$ captures the partially-observed map and the robot's pose at time $t$.
The robot is equipped with a planar laser scanner, and so it is capable of reliably mapping its surroundings in places it has already seen and determining its location with high precision, a subset of POMDPs recently coined as a \emph{Locally} Observable Markov Decision Process~\cite{merlin2020locally_lomdp}.
Robot behavior is guided by a policy $\pi$ that specifies the action the robot should take from the belief $b_t$, stored as a partial occupancy map $m_t$ and the set of visual observations collected by the robot.
During deployment, performance is measured as the average distance traveled to reach the goal across a number of trials; a trial is a single traversal from start to goal in a previously-unseen map.

\subsection{Policy Selection during Deployment}
We consider that the robot has access to a set of policies $\mathcal{P} = \{\pi_1, \pi_2, \cdots, \pi_N\}$, and its objective is to pick the policy that minimizes the expected cost during deployment,
\begin{equation} \label{eq:policy_selection}
    \pi^{*} = \argmin_{\pi \in \mathcal{P}} \mathbb{E}[C(\pi)]
\end{equation}
where $\mathbb{E}[C(\pi)]$ is the expected cost of policy $C(\pi)$ during deployment.
However, in general we will not have direct access to the expected cost and must instead estimate it during deployment via execution of multiple trials.

\subsection{Limits of Black Box Policy Selection for Planning in a Partial Map}
Black box policy selection methods---e.g., many bandit algorithms---aim to select policies by balancing minimization of the average cost observed so far and exploration, so as to occasionally select policies that have the potential to improve performance.
For trial $k+1$, the Upper Confidence Bound (UCB)\footnote{Eq.~\eqref{eq:ucb_bandit} uses a lower confidence bound (LCB) instead of a UCB, since our performance estimates are represented as \emph{costs} instead of \emph{rewards} and is therefore minimized.
As such, we use the more common term UCB to mean the approach in general rather than the upper bound itself.}~\cite{ucb_lai1985asymptotically, sutton2018reinforcement} bandit algorithm selects the next policy $\pi^{\text{\tiny(k+1)}}$ according to
\begin{equation} \label{eq:ucb_bandit}
    \pi^{\text{\tiny(k+1)}} = \argmin_{\pi \in \mathcal{P}} \Bigg[\,\, \bar{C}_k(\pi) - c\sqrt{\frac{\ln{k}}{n_k(\pi)}}\,\, \Bigg] 
\end{equation}
where $\bar{C}_k(\pi)$ is the average cost over trials 1--$k$ in which policy $\pi$ was selected, $n_k(\pi)$ is the number of times policy $\pi$ was selected (up to trial $k$), and $c>0$ is a parameter controlling the balance between exploration and exploitation.

However, for navigation under uncertainty, each trial is expensive to execute, and the results of these trials depend on the environment map and so often have high variance.
Policy selection via Eq.~\eqref{eq:ucb_bandit} is therefore often problematically slow to converge, limiting the utility of this approach in practice.
Narrowing the confidence bounds on the expected performance of each policy requires deploying them, often multiple times, and therefore the robot would potentially need to go through repeated attempts of poor performing behavior before such policies can be ruled out.
Instead, we aim to improve the data efficiency of selection via a \emph{white-box selection strategy} that uses offline alt-policy replay to compute bounds on how well each policy \emph{could have done} in the trials so far, and uses these bounds to constrain selection.

\section{Overview: Data-Efficient Policy Selection via Offline Alt-Policy Replay}\label{sec:overview}
If we are to accelerate policy selection, we must be able to quickly tighten the bounds on expected performance for each policy, prioritizing selection of the most promising policies with fewer trials.
As such, we seek to constrain policy selection by determining how well an alternative policy \emph{could have performed} if it had instead been in charge.
While we cannot re-deploy the robot to repeat the same trial with another policy, information collected during the trial can be used to scrutinize alternative behaviors and determine a lower bound on their performance even without deployment.

We seek to use \emph{deployment-time offline alt-policy replay} to compute a lower bound on the performance of policies that did not control the robot's behavior during the trial.
Using the information collected by policy $\pi$ during its trial, we can replay how every other alternative policy $\pi' \in \mathcal{P}\setminus \pi$ could have performed.
Our offline alt-policy replay makes optimistic assumptions about space not seen during the navigation trial (i.e., that the robot will only spend a minimal amount of time in unseen space), allowing us to compute a strict lower bound on the mean performance of each policy so far (see Sec.~\ref{sec:approx_lb},~\ref{sec:final_lb}).
We use this lower bound on the mean $\bar{C}^\text{\tiny{}lb}$ to constrain UCB bandit selection, so that policies that could not have improved performance are not selected.
Our \emph{Constrained} UCB Bandit algorithm selects according to
\begin{equation}\label{eq:constrained_ucb}
    \!\!\pi^{\text{\tiny(k+1)}} = \argmin_{\pi \in \mathcal{P}} \Bigg[\max \Bigg({\bar{C}^{\text{\tiny{lb}}}_k(\pi),  \bar{C}_k(\pi) - c\sqrt{\frac{\ln{k}}{n_k(\pi)}} }\,\Bigg)\Bigg] 
\end{equation}
Intuitively, Eq.~\eqref{eq:constrained_ucb} always picks the tighter (i.e., higher) of the lower bounds between (i) $\bar{C}^{\text{\tiny{lb}}}_k$, the lower bound computed via offline alt-policy replay, and (ii) bound computed by UCB algorithm as shown in Eq.~\eqref{eq:ucb_bandit} and then minimizes over this bound to pick a policy for next trial.

\section{Offline Policy Replay Requires a Planning Approach Robust to Vantage Point Change}
Scrutinizing alternative behavior to determine the lower bound on cost $\text{C}^{\text{\tiny{lb}}}_k(\pi)$ needed for selection via our constrained UCB bandit algorithm, Eq.~\eqref{eq:constrained_ucb}, requires that we can perform offline alt-policy replay of robot behavior under an alternative policy without actually deploying the robot.
In general, replaying a policy offline requires an ability to generate observations from poses the robot may not have visited, which for many learning-informed planning strategies in this domain will not accurately reflect how the policy would have behaved if it had been in control of the robot.
Many approaches to vision-informed navigation under uncertainty, particularly those relying on deep reinforcement learning~\cite{kulhanek2019vision, mirowski2016learning}, require observations (images) from poses and vantage points not visited during the original trial and can be brittle to even small changes~\cite{henderson2018rlmatters}, and so replay of such policies is unlikely to yield an accurate lower bound on cost.
As such, we instead require an approach to planning that is robust to changes in viewpoint and a kind of policy that is robust to minor changes in robot pose and corresponding observations, and can reliably reach the goal even in environments where learning informs poor behavior.

\section{Preliminaries: The Learning over Subgoals Model-Based Planning Abstraction} \label{sec:lsp}
It is a key insight that the \emph{Learning over Subgoals Planning} (LSP) of Stein et al.~\cite{stein2018learning} is well-suited for offline alt-policy replay while being suitable for long horizon planning in partially-mapped environments.
LSP is a high level planning framework in which navigation in a partially-mapped environment is formulated as a Partially Observable Markov Decision Process (POMDP).
In this abstraction, \emph{subgoals} represent the robot's high-level actions of navigating to a frontier: i.e. a boundary between free and unseen region in the map.
Each action $a_t$ corresponding to a subgoal has three properties associated with it: likelihood that the subgoal leads to the goal $P_S(a_t)$, expected cost of reaching the goal via the subgoal $R_S(a_t)$, and expected cost of failure or exploration in case the subgoal doesn't lead to the goal $R_E(a_t)$.
A neural network, parameterized by $\theta$, is trained via supervised learning to estimate the subgoal properties $\mathcal{R_{\theta}} = \{P_{S,\theta}, R_{S,\theta}, R_{E,\theta} \}$ based on image observations centered at the subgoal.
A factored form of Bellman equation shown in Eq.~\eqref{eq:bellman_lsp} is used to calculate the expected cost of each action,
\begin{multline}\label{eq:bellman_lsp}
  Q_{\theta}(b_t, a_t) = D(b_t, a_t) + P_{S,\theta}(a_t)R_{S,\theta}(a_t) \\
    + (1 - P_{S,\theta}(a_t)) \left[ R_{E,\theta}(a_t) + \min_{a \in \mathcal{A}(b_t) \backslash a_t} Q_{\theta}({b}_t, a) \right] 
\end{multline}
where $D(b_t, a_t)$ is the known cost of navigating to a frontier.
The robot's policy is then to minimize expected cost:
\begin{equation}\label{eq:lsp_policy}
    \pi_{\theta}(b_t) = \argmin_{a \in \mathcal{A}(b_t)} Q_{\theta}(b_t, a)
\end{equation}

Since learning is used only to make robot-pose-agnostic predictions of subgoal properties using panoramic image oriented so as to face the subgoal, LSP is robust to changes in the robot’s location and corresponding image observations,  and is therefore suitable for offline alt-policy replay.

\subsection{Planning via Learning over Subgoals Planning}\label{sec:lsp:p-loop}
At each time step, the robot selects a high-level action $a_t$ via Eq.~\eqref{eq:bellman_lsp} that specifies the subgoal towards which the robot will navigate.
Upon selecting a high-level action, the robot computes a cost grid over the observed map via A$^{\!*}$~\cite{hart1968formal_astar} and selects a low-level (short-horizon) motion primitive that makes the most progress towards the subgoal.
The robot (i) executes this primitive action, (ii) updates the partial map via laser scanner observations, (iii) recomputes the subgoals (and, if necessary the subgoal properties $P_S$, $R_S$, and $R_E$ from panoramic images collected onboard the robot), and finally (iv) replans via Eq.~\eqref{eq:bellman_lsp}.
As planning via LSP continually seeks out unexplored space until the goal is reached, it is guaranteed to reach the goal if a feasible path exists, even when learning informs poor behavior.

\subsection{Network Architecture and Training}\label{sec:lsp:network}
The neural network takes as input a $128 \times 512$ RGB panoramic image centered on a subgoal, the relative subgoal location and the relative goal location.
Our neural network architecture and training procedure closely follow that of Bradley et al.~\cite{bradley2021learning}.
The input image is encoded by first passing through 4 convolutional layers and then concatenated with the features representing the relative locations of subgoal and goal.
These concatenated features go through 9 convolutional layers and then 5 fully connected layers to output 3 subgoal properties: $P_S$, $R_S$, and $R_E$.

To generate training data, we conduct hundreds of trials in previously unseen maps in which the robot navigates from start to goal via a non-learned heuristic-driven planner, collecting images of the environment at each step.
The target labels for $P_S$, $R_S$, and $R_E$ corresponding to these images for observed subgoals are computed using the underlying known map.
Labels for $P_S$ correspond to whether or not a path to goal exists via a subgoal.
Labels for the costs correspond to the travel distance to reach the goal through unknown space if the goal can be reached via a subgoal (for $R_S$) and to a heuristic cost approximating the distance robot would need to travel before meeting a dead end if the goal cannot be reached via a subgoal (for $R_E$).
Further details on data collection and training can be found in~\cite{stein2018learning,bradley2021learning}.

\section{Approach: Offline Alt-Policy Replay via Learning over Subgoals} \label{sec:offline_replay}

Our approach leverages information collected during a trial to perform offline replay of policies.
During offline alt-policy replay, we make optimistic or simply-connected assumptions about the unseen space to compute an approximate lower bound cost for these policies which is then used to constrain UCB bandit to accelerate policy selection.

\subsection{Information Collection during a Trial}

During a trial (a single navigation from start to goal), we record information needed for the offline replay of other policies.
For a trial $k$, the policy $\pi^{\text{\tiny(k)}}$ determines behavior, producing a \emph{record} $\mathcal{Z}_k$ that includes 
a list of tuples $(X_t, I_t)$ where $X_t = (x_t, y_t, \phi_t)$ is the robot pose at time $t$ and $I_t$ is a panoramic image collected by the robot at time $t$.
The record $\mathcal{Z}_k$ also includes the partially-known occupancy grid map $m_\text{final}$ obtained after reaching the goal.
The record $\mathcal{Z}_k = \{ (X_t, I_t)_{\tiny t \in 1\cdots M}, m_\text{final} \}$ is then used to perform offline replay of robot behavior guided by every other policy.

\subsection{Offline Alt-Policy Replay Overview}

Upon completion of a trial (e.g., trial $k$), the robot seeks to replay behavior of all other LSP-policies $\pi' \in \mathcal{P}\setminus \pi^\text{\tiny{}(k)}$ using the record $\mathcal{Z}_k$ it collected during the trial.
Offline alt-policy replay proceeds similarly to online planning via LSP, as described in Sec.~\ref{sec:lsp:p-loop}:
at every replay time step $\tau$, the robot updates its map using a simulated laser scan, computed by ray-casting into the 2D partial map $m_\text{final}$, and subsequently recomputes the available subgoals, each corresponding to a boundary between free and unknown space.

To compute the subgoal properties $P_S$, $R_S$, $R_E$ associated with exploration into a particular region of unseen space, the robot requires panoramic images to be fed into its neural network (Sec.~\ref{sec:lsp:network}).
Though images will not be available at every point in the environment during offline alt-policy replay, subgoal properties are instead estimated from nearby images stored in $\mathcal{Z}_k$.
Specifically, we find the pose $X_\text{nearest}$ from $\mathcal{Z}_k$ nearest to the current (replay) pose $X_\tau$ that can see the subgoal of interest and then use its associated image $I_\text{nearest}$ to estimate the subgoal properties.
Using the estimated subgoal properties, the robot selects a high-level action via Eq.~\eqref{eq:bellman_lsp}, executes a low-level motion primitive to navigate towards the selected subgoal-action, and this process repeats.
Critically, if we replay the deployed policy $\pi^\text{\tiny{}(k)}$ using its own record $\mathcal{Z}_k$, we recover its behavior exactly.

\subsection{Computing an Approximate Lower Bound Cost} \label{sec:approx_lb}
\begin{figure}[t]
    \centering
    \vspace{0.6em}
    \includegraphics[width=8.6cm]{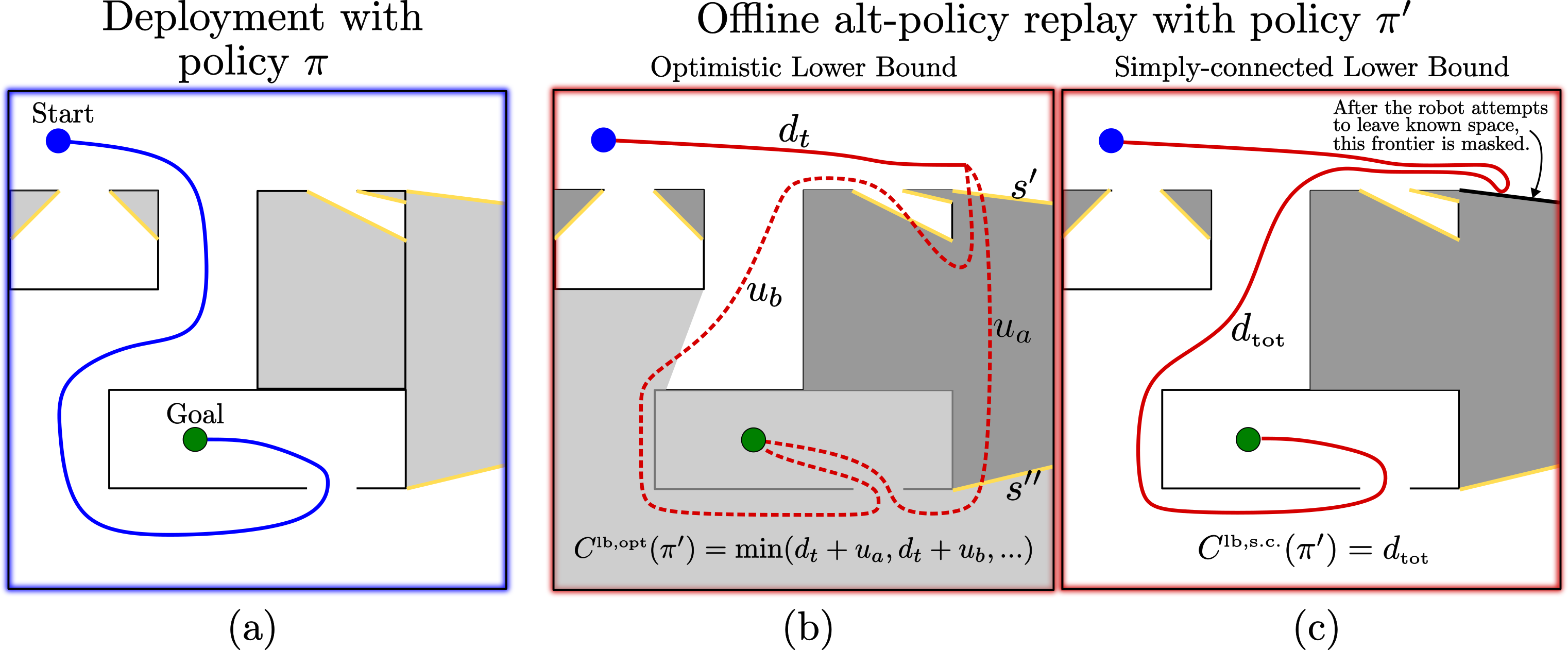}
    \vspace{-2.0em}
    \caption{\textbf{Lower Bound Cost Approximation}: (a) Policy $\pi$ guides the robot in a trial. (b) During offline alt-policy replay, policy $\pi'$ attempts to leave known space via subgoal $s'$ to try to reach goal with proposed path $u_a$. The minimum of such paths obtained during replay gives the optimistic lower bound of policy $\pi'$. (c) With simply-connected assumption, the lower bound is the net distance travelled under policy $\pi'$ during offline replay.}
    \label{fig:lower_bound_cost}
    \vspace{-1.5em}
\end{figure}

As the robot navigates to goal during offline alt-policy replay, it might attempt to enter space not seen by the robot planning with policy $\pi^{\text{\tiny(k)}}$ and thus unknown in the partial map $m_\text{final}$.
Since this space is not known, we cannot know precisely what will happen in this unseen space and so we instead make conservative assumptions about what \emph{could} happen: we assume that the robot either (i) reaches the goal in the shortest possible distance through unseen space or (ii) is immediately turned around and must pursue a different route to the goal.
Based on what we know about unseen space---e.g., whether or not the environment is known to be simply-connected or the likelihood that there exists an undiscovered shortcut to the goal---we can make different assumptions and use them to compute lower bounds on the cost.
Here (and visualized in Fig.~\ref{fig:lower_bound_cost}) we discuss these different assumptions, their implications, and when each is appropriate.

\textbf{Optimistic Cost Lower Bound}\quad{}
If we have no prior information about the structure of unseen space, we make an \emph{optimistic} assumption about unseen space: that all unseen space is free and could therefore provide a potential route to the goal.
Under this assumption, whenever the actions from offline-replayed policy leads the robot to leave known space (via a subgoal $s'$), we assume the robot \emph{could} have reached the goal via the shortest possible path in $m_\text{final}$ that passes through $s'$.
Fig. \ref{fig:lower_bound_cost}(b) shows two examples of such alternative paths.
Upon attempting to leave known space, we compute the optimistic path cost, mask that particular frontier (so that the robot must continue exploration through known space), and proceed with navigation.
We generate such optimistic path costs every time the offline-replayed robot attempts to leave known space.
The shortest of these paths is the optimistic lower bound cost $C^{\text{\tiny{}lb,opt}}_k(\pi')$ for a replayed policy $\pi'$ in trial $k$.

\textbf{Simply-Connected Cost Lower Bound}
If we were to know in advance that the environments were simply-connected, we could come up with a tighter lower bound on the cost, since the shortest path to the goal should exist within known space in $m_\text{final}$ and routes that leave observed space will not reach the goal.
During offline-replayed navigation, whenever the robot attempts to leave known space, the boundary through which it aims to leave is masked as an obstacle before it is allowed to do so, forcing it to turn back and seek an alternate route to the goal, Fig.~\ref{fig:lower_bound_cost}(c).
The total distance traveled to reach the goal is the simply-connected lower bound cost $C^\text{\tiny{}lb,s.c.}_k(\pi')$ for a replayed policy $\pi'$.

\textbf{Weighted Approximate Cost Lower Bound}
In many environments, the optimistic lower bound is not particularly tight, and so does not help to accelerate selection, yet if the environment is not known to be simply connected, the simply connected lower bound may be too high and not a lower bound on cost.
Often, we can use our prior understanding of an environment to determine the likelihood that alternative shortcuts to the goal could exist in space the robot did not see during its trial.
We introduce a parameter $p_{\text{\tiny{}short}}$ that denotes the likelihood of the existence of a shorter path to goal.
This parameter is used to compute a third \emph{approximate} lower bound, defined as the weighted sum of the optimistic and simply-connected bounds:
\begin{equation}\label{eq:weighted_lb}
C_{k,p}^\text{\tiny{}lb,wgt}(\pi') = 
p_\text{\tiny{}short}C^\text{\tiny{}lb,opt}_k(\pi') + 
(1-p_\text{\tiny{}short})C^\text{\tiny{}lb,s.c.}_k(\pi')
\end{equation}

While the weighted approximate lower bound cost may not strictly be a lower bound, and so may violate guarantees of asymptotic regret bounds afforded by UCB bandit selection, we will show that this approximate lower bound cost helps to achieve good empirical performance (Sec.~\ref{sec:office_results}).

\subsection{Combining Observed and Replayed Lower Bound Costs} \label{sec:final_lb}
As trials proceed, we compute an approximate lower bound on the mean $\bar{C}^\text{\tiny{}lb}$, that combines both averaged performance $\bar{C}_k(\pi)$ of policy $\pi$ until trial $k$ and the mean replayed lower bound $\bar{C}^\text{\tiny{}lb,rep}$, the average cost computed from offline alt-policy replay via one of the lower bounds defined in Sec.~\ref{sec:approx_lb}:
\begin{equation} \label{eq:final_cost_lb}
    \bar{C}^\text{\tiny{lb}}_k(\pi) = \frac{
    n_k(\pi)\bar{C}_k(\pi) + n_k^\text{\tiny{}rep}(\pi) \bar{C}^\text{\tiny{}lb,rep}}{
    n_k(\pi) + n_k^\text{\tiny{}rep}(\pi)}
\end{equation}
where $n_k(\pi)$ and $n_k^\text{\tiny{}rep}(\pi)$ are the number of times each policy has been deployed and offline-replayed, respectively.

Constrained policy selection based on modified UCB bandit formulation discussed in Sec.~\ref{sec:overview} leverages this lower bound $\bar{C}^{\text{\tiny{lb}}}_k(\pi)$ to quickly identify the best-performing policy.

\section{Experimental Results} \label{sec:experiments}
\begin{figure}
    \centering
    \vspace{0.5em}
    \includegraphics[width=8.6cm]{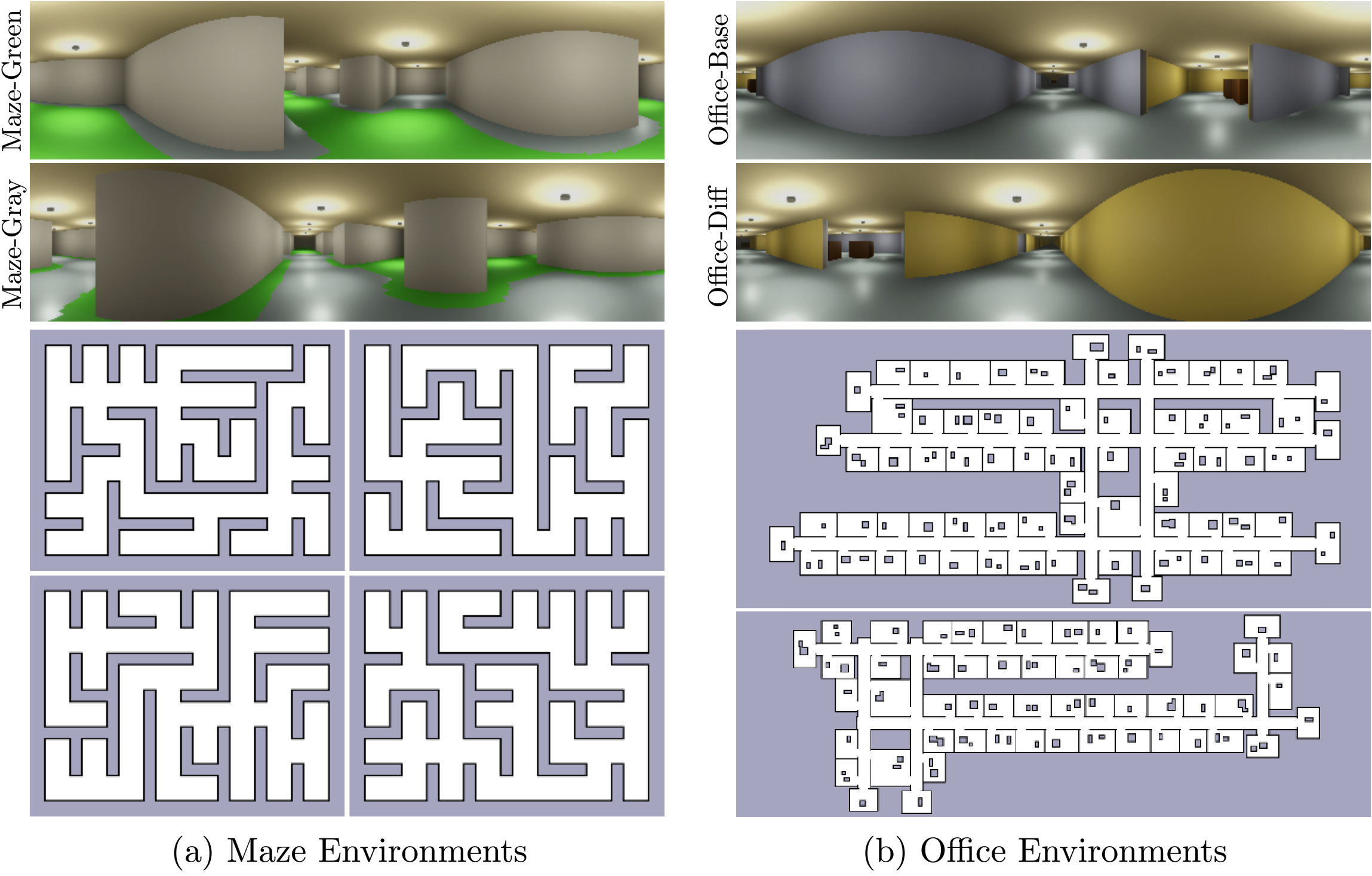}
    \vspace{-1.7em}
    \caption{\textbf{The Simulated Environments}.\quad{} Robot-view panoramic images from simulation environments (top two rows) and samples of maps from (a) maze (b) office-like environments. All our experiments are conducted in simulated environments rendered using the Unity game engine.}
    \vspace{-1.5em}
    \label{fig:environments}
\end{figure}

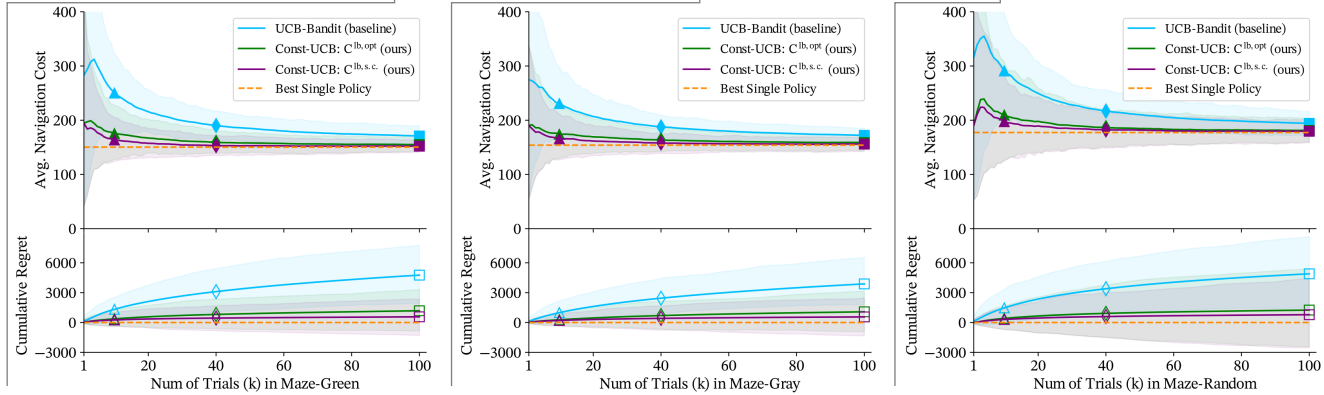
\begin{figure*}
    \centering
    \vspace{0.5em}
    \footnotesize

    \begin{tabular}{llr@{\hspace{1.5pt}}r@{\hspace{1.5pt}}rr@{\hspace{1.5pt}}r@{\hspace{1.5pt}}rr@{\hspace{1.5pt}}r@{\hspace{1.5pt}}r}
    \toprule
                & Deployment Environment\tikzmark{mazept3} & \multicolumn{3}{c}{Maze-Green}\tikzmark{mazept1} & \multicolumn{3}{c}{Maze-Gray}\tikzmark{mazept2} & \multicolumn{3}{c}{Maze-Random}\tikzmark{mazept4} \\
    \cmidrule{1-11}
                & Num of Trials ($k$) & \multicolumn{1}{c}{$k=10$}     & \multicolumn{1}{c}{$k=40$}     & \multicolumn{1}{c}{$k=100$}    
                  & \multicolumn{1}{c}{$k=10$}     & \multicolumn{1}{c}{$k=40$}     & \multicolumn{1}{c}{$k=100$}   
                  & \multicolumn{1}{c}{$k=10$}     & \multicolumn{1}{c}{$k=40$}     & \multicolumn{1}{c}{$k=100$}    \\
\midrule
\multirow{4}{*}{\rotatebox[origin=c]{90}{\shortstack{Average\\ Nav. Cost\\ (mean)}}}

& UCB-Bandit (baseline)      
                & 250.04\textcolor{dbl}{$\blacktriangle$}   & 190.08\textcolor{dbl}{$\blacklozenge$}  & 170.88\textcolor{dbl}{$\blacksquare$}
                & 230.66\textcolor{dbl}{$\blacktriangle$}  & 187.71\textcolor{dbl}{$\blacklozenge$}  & 172.10\textcolor{dbl}{$\blacksquare$}
                & 290.89\textcolor{dbl}{$\blacktriangle$}   & 216.87\textcolor{dbl}{$\blacklozenge$}  & 194.20\textcolor{dbl}{$\blacksquare$}  \\
& Const-UCB:$C^\text{\tiny{}lb,opt}$ (ours) 
                & 175.38\textcolor{grn}{$\blacktriangle$}   & 159.14\textcolor{grn}{$\blacklozenge$}  & 155.08\textcolor{grn}{$\blacksquare$}  
                & 173.96\textcolor{grn}{$\blacktriangle$}  & 163.50\textcolor{grn}{$\blacklozenge$}  & 158.81\textcolor{grn}{$\blacksquare$} 
                & 209.44\textcolor{grn}{$\blacktriangle$}   & 186.21\textcolor{grn}{$\blacklozenge$}  & 181.20\textcolor{grn}{$\blacksquare$}  \\
& Const-UCB:$C^\text{\tiny{}lb,s.c.}$ (ours) 
                & \textbf{163.56}\textcolor{prp}{$\blacktriangle$}   & \textbf{153.12}\textcolor{prp}{$\blacklozenge$}  & \textbf{152.13}\textcolor{prp}{$\blacksquare$}  
                & \textbf{165.62}\textcolor{prp}{$\blacktriangle$}  & \textbf{158.06}\textcolor{prp}{$\blacklozenge$}  & \textbf{155.94}\textcolor{prp}{$\blacksquare$} 
                & \textbf{197.30}\textcolor{prp}{$\blacktriangle$}   & \textbf{182.14}\textcolor{prp}{$\blacklozenge$}  & \textbf{179.66}\textcolor{prp}{$\blacksquare$} \\
    \addlinespace[-0.33em]
    \cmidrule{2-11}
    \addlinespace[-0.3em]
& Best Single Policy\tablefootnote{For each environment, ``Best Single Policy'' refers to the policy which incurs minimum cost when deployed in that environment. The costs incurred by Best Single Policy are underlined in Table \ref{tab:maze_base_costs} and \ref{tab:office_base_costs}.\label{note:best_policy}} 
& 150.39\textcolor{white}{$\blacktriangle$} & 150.39\textcolor{white}{$\blacklozenge$} & 150.39\textcolor{white}{$\blacksquare$} 
                & 154.03\textcolor{white}{$\blacktriangle$} & 154.03\textcolor{white}{$\blacklozenge$} & 154.03\textcolor{white}{$\blacksquare$} 
                & 177.31\textcolor{white}{$\blacktriangle$} & 177.31\textcolor{white}{$\blacklozenge$} & 177.31\textcolor{white}{$\blacksquare$}  \\

    \addlinespace[-0.3em]
    \midrule

\multirow{3}{*}{\rotatebox[origin=c]{90}{\shortstack{Cumul.\\ Regret\\ (mean)}}}

& UCB-Bandit (baseline)      
                & 1327.0\textcolor{dbl}{$\pmb{\vartriangle}$}   & 3105.9\textcolor{dbl}{$\pmb{\lozenge}$}  & 4761.4\textcolor{dbl}{$\pmb{\square}$}
                & 994.1\textcolor{dbl}{$\pmb{\vartriangle}$}  & 2444.8\textcolor{dbl}{$\pmb{\lozenge}$}  & 3879.4\textcolor{dbl}{$\pmb{\square}$}
                & 1472.0\textcolor{dbl}{$\pmb{\vartriangle}$}   & 3391.9\textcolor{dbl}{$\pmb{\lozenge}$}  & 4892.1\textcolor{dbl}{$\pmb{\square}$}  \\
& Const-UCB:$C^\text{\tiny{}lb,opt}$ (ours) 
                & 369.1\textcolor{grn}{$\pmb{\vartriangle}$}   & 813.4\textcolor{grn}{$\pmb{\lozenge}$}  & 1180.5\textcolor{grn}{$\pmb{\square}$}  
                & 278.2\textcolor{grn}{$\pmb{\vartriangle}$}  & 698.9\textcolor{grn}{$\pmb{\lozenge}$}  & 1076.9\textcolor{grn}{$\pmb{\square}$} 
                & 417.8\textcolor{grn}{$\pmb{\vartriangle}$}   & 918.6\textcolor{grn}{$\pmb{\lozenge}$}  & 1249.1\textcolor{grn}{$\pmb{\square}$}  \\
& Const-UCB:$C^\text{\tiny{}lb,s.c.}$ (ours) 
                & \textbf{259.2}\textcolor{prp}{$\pmb{\vartriangle}$}   & \textbf{446.6}\textcolor{prp}{$\pmb{\lozenge}$}  & \textbf{573.3}\tikzmark{end}{\textcolor{prp}{$\pmb{\square}$}} 
                & \textbf{208.4}\textcolor{prp}{$\pmb{\vartriangle}$}  & \textbf{419.1}\textcolor{prp}{$\pmb{\lozenge}$}  & \textbf{565.0}\textcolor{prp}{$\pmb{\square}$} 
                & \textbf{309.5}\textcolor{prp}{$\pmb{\vartriangle}$}   & \textbf{600.6}\textcolor{prp}{$\pmb{\lozenge}$}  & \textbf{786.9}\textcolor{prp}{$\pmb{\square}$} \\
    \end{tabular}

    \includegraphics[width=17.7cm]{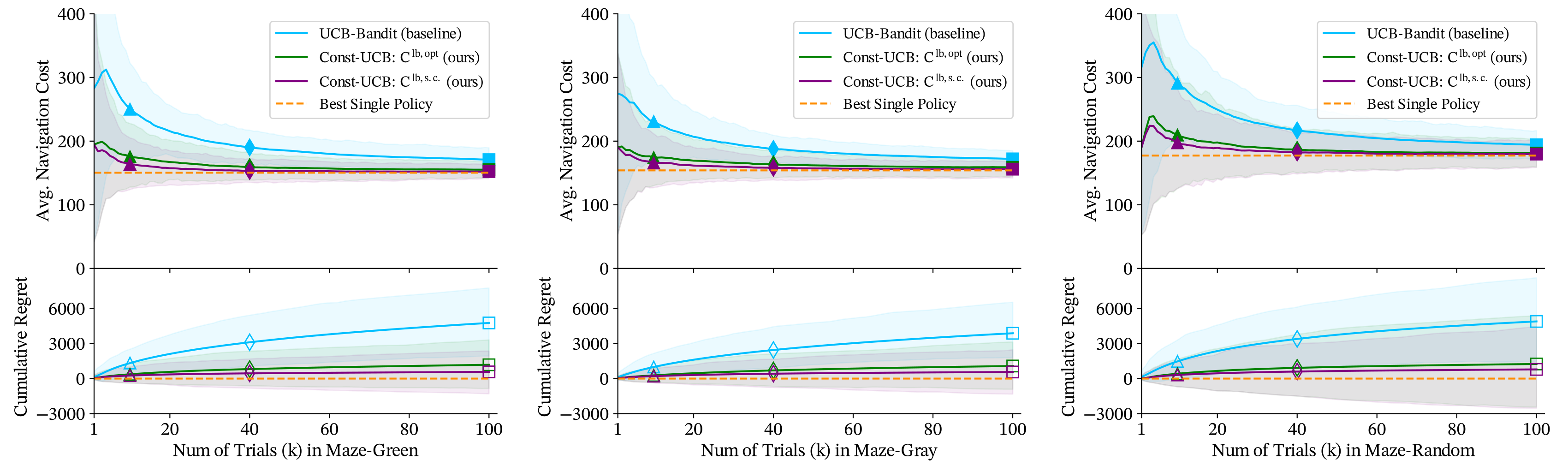}
    \vspace{-1.5em}
    \caption{\textbf{Average Navigation Cost (mean) and Cumulative Regret (mean) for deployments in maze environments, Fig.~\ref{fig:environments}(a).}\quad{} Each deployment consists of 100 randomized navigation trials, each in a previously unseen maze. Mean cost and regret are computed across 200 randomized deployments. For our approach Const-UCB, we show results with optimistic $C^\text{\tiny{}lb,opt}$ and simply connected $C^\text{\tiny{}lb,s.c.}$ lower bounds as discussed in Sec.~\ref{sec:approx_lb}. The solid lines denote the mean, and the shaded regions show 10\textsuperscript{th} to 90\textsuperscript{th} percentile. The symbols: triangle, diamond and square denote average cost (filled) and cumulative regret (unfilled) at 10\textsuperscript{th}, 40\textsuperscript{th} and 100\textsuperscript{th} trial respectively in both the table and the plot for each environment.}
    \vspace{-1.5em}
    \label{fig:results_maze}

    \begin{tikzpicture}[remember picture,overlay,color=gray]
        \draw[solid, line width=0.5pt] (pic cs:mazept1) ++(0.15, 0.33) -- ++(0, -3.46) -- ++(-3.3, 0) -- ++ (0,-5.1);
        \draw[solid, line width=0.5pt] (pic cs:mazept2) ++(0.15, 0.33) -- ++(0, -3.46) -- ++(-1.4, 0) -- ++ (0,-5.1);
        \draw[solid, line width=0.5pt] (pic cs:mazept3) ++(0.2, 0.33) -- ++(0, -3.46) -- ++(-5.15, 0) -- ++ (0,-5.1);
    \end{tikzpicture}

\end{figure*}

We perform experiments in simulated maze and office-like environments (Fig.~\ref{fig:environments}), evaluating our constrained policy selection approach, Const-UCB, Eq.~\eqref{eq:constrained_ucb} against baseline UCB bandit selection, Eq.~\eqref{eq:ucb_bandit}.
We conduct 200 deployments, each consists of 100 navigation trials, each in a distinct, procedurally-generated map not yet seen by the robot.
At the outset of each deployment, the robot starts with a randomly selected policy.
The policy for subsequent trials is selected via our Const-UCB approach or the UCB bandit.
Both our Const-UCB approach and the baseline UCB bandit approach use an exploration parameter $c=100$ in all experiments, empirically chosen on a held out test set to achieve good performance on the baseline UCB bandit approach.
We note that the results are not particularly sensitive to changes in $c$.
We additionally show results of deploying each policy individually in the absence of policy selection to illustrate the necessity of deployment-time selection for good performance across environments.

\subsection{Maze-centric Results}\label{sec:maze_results}

Our simulated maze environments consist of randomly generated simply-connected mazes in which the robot needs to navigate from start to unseen goal.
Each generated map is unique with randomized start and goal poses.
To study the versatility and effectiveness of our approach, we design three variations of the maze environments and train an LSP-based policy (Sec.~\ref{sec:lsp}) in each.
Shown in Fig.~\ref{fig:environments}, the three maze environment variations are as follows:

\begin{table}
    \centering
    \caption{Average navigation cost for each policy in maze-centric environments without policy selection}
    \vspace{-0.5em}
    \begin{tabular}{lccc}
    \toprule
            & Maze-Green & Maze-Gray & Maze-Random \\
    \midrule
    Non-learned     & 206.05     & 194.37    & \underline{177.31}      \\
    LSP-Maze-Green  & \underline{150.39}     & 483.20    & 557.99      \\
    LSP-Maze-Gray & 618.87     & \underline{154.03}    & 418.98      \\
    LSP-Maze-Random & 231.71     & 238.22    & 180.23      \\
    \bottomrule
    \vspace{-2.5em}
    \end{tabular}
    \label{tab:maze_base_costs}
\end{table}

\begin{LaTeXdescription}
    \item[Maze-Green] The floor is generally gray, but a green path on the ground connects the start to goal. The green color is thus a signal an experienced agent should recognize as leading towards the unseen goal.
    \item[Maze-Gray] Similar to the Maze-Green environment, yet the color of the floor and path are flipped: the floor is green with a gray path to the goal. Thus, this environment should mislead policies trained in Maze-Green.
    \item[Maze-Random] The green path on gray floor is placed randomly and is not a reliable route to goal.
\end{LaTeXdescription}

We train a LSP-based policy in each---yielding LSP-Maze-Green, LSP-Maze-Gray, and LSP-Maze-Random---following the procedure in Sec.~\ref{sec:lsp:network}.
Each LSP-based policy is trained on 500 randomly generated mazes.
Each maze map during a navigation trial is also distinct and is not a part of the training set.
We also deploy a non-learned optimistic baseline policy that our approach can select; the non-learned policy, instead of using a neural network, uses optimistic heuristics about the environment to compute the subgoal properties, namely that each subgoal could lead to the unseen goal: i.e., $P_S=1$.

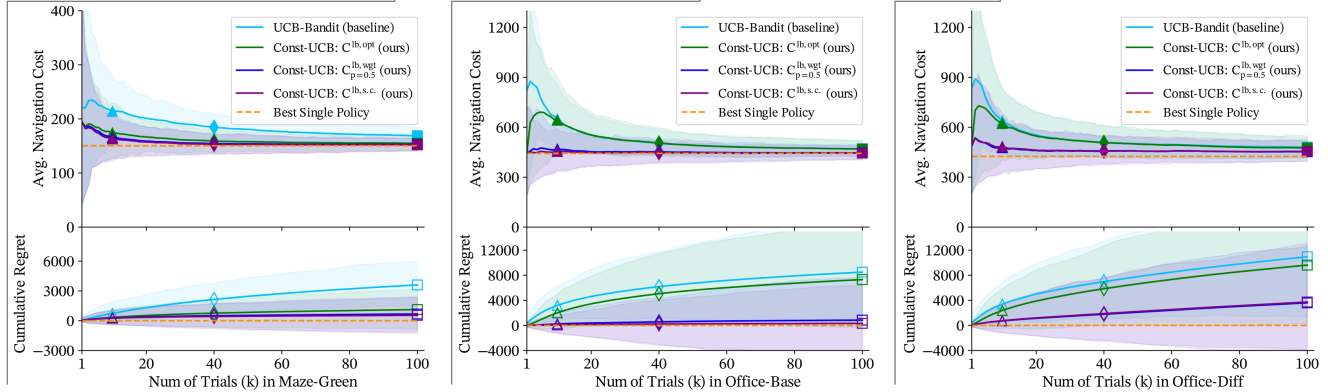
\begin{figure*}[t]
    \centering
    \vspace{0.5em}
    \footnotesize

    \begin{tabular}{llr@{\hspace{1.5pt}}r@{\hspace{1.5pt}}rr@{\hspace{1.5pt}}r@{\hspace{1.5pt}}rr@{\hspace{1.5pt}}r@{\hspace{1.5pt}}r}
    \toprule
                & Deployment Environment\tikzmark{officept3} & \multicolumn{3}{c}{Maze-Green}\tikzmark{officept1} & \multicolumn{3}{c}{Office-Base}\tikzmark{officept2} & \multicolumn{3}{c}{Office-Diff}\tikzmark{officept4} \\
    \cmidrule{1-11}
                & Num of Trials ($k$) & \multicolumn{1}{c}{$k=10$}     & \multicolumn{1}{c}{$k=40$}     & \multicolumn{1}{c}{$k=100$}    
                  & \multicolumn{1}{c}{$k=10$}     & \multicolumn{1}{c}{$k=40$}     & \multicolumn{1}{c}{$k=100$}   
                  & \multicolumn{1}{c}{$k=10$}     & \multicolumn{1}{c}{$k=40$}     & \multicolumn{1}{c}{$k=100$}    \\
    \midrule
\multirow{5}{*}{\rotatebox[origin=c]{90}{\shortstack{Average\\ Navigation\\ Cost (mean)}}}

& UCB-Bandit (baseline)      & 213.90\textcolor{dbl}{$\blacktriangle$}   & 184.62\textcolor{dbl}{$\blacklozenge$}  & 168.87\textcolor{dbl}{$\blacksquare$}
                & 648.02\textcolor{dbl}{$\blacktriangle$}  & 503.31\textcolor{dbl}{$\blacklozenge$}  & 470.43\textcolor{dbl}{$\blacksquare$}
                & 638.02\textcolor{dbl}{$\blacktriangle$}   & 508.72\textcolor{dbl}{$\blacklozenge$}  & 480.22\textcolor{dbl}{$\blacksquare$}  \\
& Const-UCB:$C^\text{\tiny{}lb,opt}$ (ours) & 174.03\textcolor{grn}{$\blacktriangle$}   & 158.88\textcolor{grn}{$\blacklozenge$}  & 155.03\textcolor{grn}{$\blacksquare$}  
                & 642.21\textcolor{grn}{$\blacktriangle$}  & 505.00\textcolor{grn}{$\blacklozenge$}  & 469.48\textcolor{grn}{$\blacksquare$}
                & 622.17\textcolor{grn}{$\blacktriangle$}   & 508.17\textcolor{grn}{$\blacklozenge$}  & 475.46\textcolor{grn}{$\blacksquare$}  \\
& Const-UCB:$C_{\text{\tiny{}p=0.5}}^\text{\tiny{}lb,wgt}$ (ours) & 165.65\textcolor{blu}{$\blacktriangle$}   & 154.23\textcolor{blu}{$\blacklozenge$}  & 152.63\textcolor{blu}{$\blacksquare$}  
                & 467.63\textcolor{blu}{$\blacktriangle$}  & 451.30\textcolor{blu}{$\blacklozenge$}  & 446.58\textcolor{blu}{$\blacksquare$} 
                & \textbf{476.77}\textcolor{blu}{$\blacktriangle$}   & \textbf{456.97}\textcolor{blu}{$\blacklozenge$}  & \textbf{452.43}\textcolor{blu}{$\blacksquare$} \\
& Const-UCB:$C^\text{\tiny{}lb,s.c.}$ (ours) & \tikzmark{hello}{\textbf{163.01}}\textcolor{prp}{$\blacktriangle$}   & \textbf{152.91}\textcolor{prp}{$\blacklozenge$}  & \textbf{152.03}\textcolor{prp}{$\blacksquare$}  
                & \textbf{454.22}\textcolor{prp}{$\blacktriangle$}  & \textbf{446.93}\textcolor{prp}{$\blacklozenge$}  & \textbf{444.77}\textcolor{prp}{$\blacksquare$} 
                & 483.54\textcolor{prp}{$\blacktriangle$}   & 458.35\textcolor{prp}{$\blacklozenge$}  & 454.27\textcolor{prp}{$\blacksquare$}  \\
    \addlinespace[-0.33em]
    \cmidrule{2-11}
    \addlinespace[-0.3em]

& Best Single Policy\tablefootnotemark{note:best_policy}
            & 150.39\textcolor{white}{$\blacktriangle$} & 150.39\textcolor{white}{$\blacklozenge$} & 150.39\textcolor{white}{$\blacksquare$} 
            & 442.95\textcolor{white}{$\blacktriangle$} & 442.95\textcolor{white}{$\blacklozenge$} & 442.95\textcolor{white}{$\blacksquare$} 
            & 424.84\textcolor{white}{$\blacktriangle$} & 424.84\textcolor{white}{$\blacklozenge$} & 424.84\textcolor{white}{$\blacksquare$}  \\

    \addlinespace[-0.3em]
    \midrule
\multirow{4}{*}{\rotatebox[origin=c]{90}{\shortstack{Cumul.\\ Regret \\ (mean)}}}

& UCB-Bandit (baseline)      & 736.7\textcolor{dbl}{$\pmb{\vartriangle}$}   & 2138.7\textcolor{dbl}{$\pmb{\lozenge}$}  & 3603.1\textcolor{dbl}{$\pmb{\square}$}
                & 3180.7\textcolor{dbl}{$\pmb{\vartriangle}$}  & 6186.5\textcolor{dbl}{$\pmb{\lozenge}$}  & 8495.4\textcolor{dbl}{$\pmb{\square}$}
                & 3389.6\textcolor{dbl}{$\pmb{\vartriangle}$}   & 6978.6\textcolor{dbl}{$\pmb{\lozenge}$}  & 10948.8\textcolor{dbl}{$\pmb{\square}$}  \\
& Const-UCB:$C^\text{\tiny{}lb,opt}$ (ours) & 332.2\textcolor{grn}{$\pmb{\vartriangle}$}   & 760.7\textcolor{grn}{$\pmb{\lozenge}$}  & 1121.3\textcolor{grn}{$\pmb{\square}$}  
                & 2003.2\textcolor{grn}{$\pmb{\vartriangle}$}  & 5053.6\textcolor{grn}{$\pmb{\lozenge}$}  & 7309.7\textcolor{grn}{$\pmb{\square}$} 
                & 2374.3\textcolor{grn}{$\pmb{\vartriangle}$}   & 5848.0\textcolor{grn}{$\pmb{\lozenge}$}  & 9615.9\textcolor{grn}{$\pmb{\square}$}  \\
& Const-UCB:$C_{\text{\tiny{}p=0.5}}^\text{\tiny{}lb,wgt}$ (ours) & 274.9\textcolor{blu}{$\pmb{\vartriangle}$}   & 504.0\textcolor{blu}{$\pmb{\lozenge}$}  & 674.8\textcolor{blu}{$\pmb{\square}$}  
                & 222.1\textcolor{blu}{$\pmb{\vartriangle}$}  & 568.5\textcolor{blu}{$\pmb{\lozenge}$}  & 847.5\textcolor{blu}{$\pmb{\square}$} 
                & \textbf{719.6}\textcolor{blu}{$\pmb{\vartriangle}$}   & \textbf{1783.2}\textcolor{blu}{$\pmb{\lozenge}$}  & \textbf{3599.2}\textcolor{blu}{$\pmb{\square}$}  \\
& Const-UCB:$C^\text{\tiny{}lb,s.c.}$ (ours) & \textbf{249.8}\textcolor{prp}{$\pmb{\vartriangle}$}   & \textbf{428.9}\textcolor{prp}{$\pmb{\lozenge}$}  & \textbf{546.4}\textcolor{prp}{$\pmb{\square}$}  
                & \textbf{74.5}\textcolor{prp}{$\pmb{\vartriangle}$}  & \textbf{230.9}\textcolor{prp}{$\pmb{\lozenge}$}  & \textbf{327.4}\textcolor{prp}{$\pmb{\square}$} 
                & 749.3\textcolor{prp}{$\pmb{\vartriangle}$}   & 1887.6\textcolor{prp}{$\pmb{\lozenge}$}  & 3737.3\textcolor{prp}{$\pmb{\square}$} \\
    \end{tabular}

    \includegraphics[width=17.7cm]{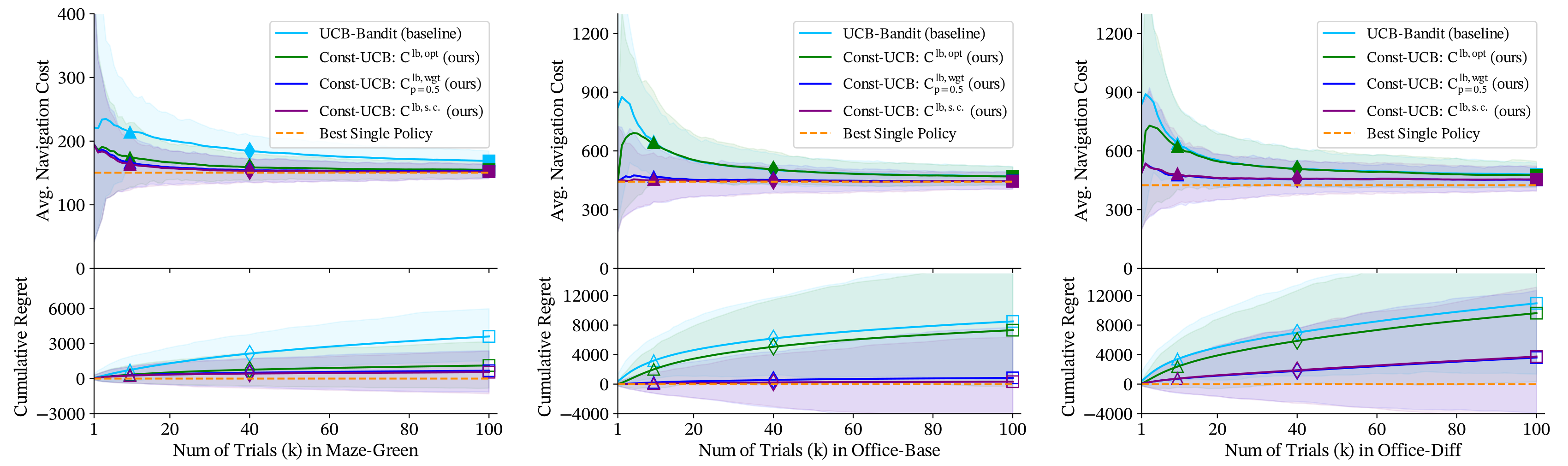}
    \vspace{-1.5em}
    \caption{\textbf{Average Navigation Cost (mean) and Cumulative Regret (mean) for deployments in office-centric environments, Fig.~\ref{fig:environments}(b).}\quad{} Each deployment consists of 100 randomized navigation trials, each in a previously unseen map. Mean cost and regret are computed across 200 randomized deployments. For our approach Const-UCB, we show results with optimistic $C^\text{\tiny{}lb,opt}$, weighted $C^\text{\tiny{}lb,wgt}$ and simply connected $C^\text{\tiny{}lb,s.c.}$ lower bounds as discussed in Sec.~\ref{sec:approx_lb}. The solid lines denote the mean, and the shaded regions show 10\textsuperscript{th} to 90\textsuperscript{th} percentile. The symbols: triangle, diamond and square denote average cost (filled) and cumulative regret (unfilled) at 10\textsuperscript{th}, 40\textsuperscript{th} and 100\textsuperscript{th} trial respectively in both the table and the plot for each environment.}
    \vspace{-1.5em}
    \label{fig:results_office}

    \begin{tikzpicture}[remember picture,overlay,color=gray]
        \draw[solid, line width=0.5pt] (pic cs:officept1) ++(0.15, 0.33) -- ++(0, -4.2) -- ++(-3.3, 0) -- ++ (0,-5.1);
        \draw[solid, line width=0.5pt] (pic cs:officept2) ++(0.15, 0.33) -- ++(0, -4.2) -- ++(-1.4, 0) -- ++ (0,-5.1);
        \draw[solid, line width=0.5pt] (pic cs:officept3) ++(0.2, 0.33) -- ++(0, -4.2) -- ++(-5.15, 0) -- ++ (0,-5.1);
    \end{tikzpicture}

\end{figure*}

We evaluate policy selection approach across 100 trials in each maze variation.
Each trial consists of a randomly generated maze not present during the training of corresponding policies.
Policy selection evaluation for each trial\protect\footnote{For computational efficiency when computing statistics, each deployment randomly samples a subset of 100 distinct evaluation scenarios from a set of 150, run in advance of model selection for each policy.} is aggregated over 200 randomized deployments to compute statistics; we compute the mean and upper and lower 10\textsuperscript{th}-percentile and show results in Fig.~\ref{fig:results_maze}.
We show average navigation cost and cumulative regret accrued until $k$\textsuperscript{th} trial for each approach.
UCB-Bandit is the baseline policy selection approach using Eq.~\eqref{eq:ucb_bandit}, and Const-UCB corresponds to our constrained policy selection approach using Eq.~\eqref{eq:constrained_ucb}.

The results in Fig.~\ref{fig:results_maze} show that in all environments, our constrained UCB (Const-UCB) bandit approach accelerates policy selection, reducing average navigation costs within far fewer trials than is possible with the UCB-Bandit alone.
Consequently, our Const-UCB approach accumulates significantly lower mean cumulative regret compared to UCB bandit: 88\% lower in Maze-Green, 85\% lower in Maze-Gray and 84\% lower in Maze-Random owing to quickly ruling out poor performing policies.
The maze environments are constructed so as to be simply-connected and selection with the simply-connected lower bound $C^\text{\tiny{}lb,s.c.}$ achieves the best performance.
Even if we did not know in advance that the environments were simply-connected, our selection procedure using the optimistic lower bound $C^\text{\tiny{}lb,opt}$ outperforms baseline UCB bandit, demonstrating the utility of our approach.

\subsection{Office-centric Results} \label{sec:office_results}

We simulate navigation in procedurally-generated office-like environments consisting of randomly generated interconnected hallways with rooms meant to look like offices with furniture-like clutter.
Each generated map is unique with randomized start and goal poses.
Fig.~\ref{fig:environments} shows example offices from our visual simulator.
We design and experiment in two variants of the office environments:

\begin{table}
    \centering
    \caption{Average navigation cost for each policy in office-centric environments without policy selection}
    \vspace{-0.5em}
    \begin{tabular}{lccc}
    \toprule
            & Maze-Green & Office-Base & Office-Diff \\
    \midrule
    Non-learned     & 206.05     & 448.60    & 476.88      \\
    LSP-Maze-Green  & \underline{150.39}     & 1667.70    & 1759.76      \\
    LSP-Office-Base & 258.76     & \underline{442.95}    & 641.88      \\
    LSP-Office-Diff & 306.45     & 942.62    & \underline{424.84}      \\
    \bottomrule
    \vspace{-2.5em}
    \end{tabular}
    \label{tab:office_base_costs}
\end{table}

\begin{LaTeXdescription}
    \item[Office-Base] The hallway walls are painted with gray while room walls are yellow.
    \item[Office-Diff] The wall colors for hallways and rooms are swapped to look visually different from Office-Base.
\end{LaTeXdescription}

Two LSP-based policies, LSP-Office-Base and LSP-Office-Diff, are trained in their corresponding environments following the procedure in Sec.~\ref{sec:lsp:network}.
Similar to maze experiments, each LSP-based policy in office environments is trained in 500 randomly generated offices while deploying on a different set of maps.
In addition to these two policies, we deploy an optimistic non-learned policy and the LSP-Maze-Green policy (see Sec.~\ref{sec:maze_results} for both), the latter of which is trained in Maze-Green, showing cross-environment policy selection and thus the flexibility of our approach.
Deployment and statistics generation follow the same procedure as described for maze experiments (Sec. \ref{sec:maze_results}).

The results in Fig.~\ref{fig:results_office} show that our constrained UCB (Const-UCB) bandit approach reduces the average navigation cost within fewer trials compared to the baseline UCB bandit approach in all of our experiments, more quickly converging closer to the costs of the best performing policy, and never under-performs the bandit.
Consequently, our Const-UCB approach accumulates significantly lower mean cumulative regret compared to UCB bandit: 85\% lower in Maze-Green (with $C^\text{\tiny{}lb,s.c.}$ bound), 96\% lower in Office-Base (with $C^\text{\tiny{}lb,s.c.}$ bound) and 67\% lower in Office-Diff (with $C_{\text{\tiny{}p=0.5}}^\text{\tiny{}lb,wgt}$ bound) owing to quickly ruling out poor performing policies.

Our office environment is not simply connected, and so only the optimistic lower bound $C^\text{\tiny{}lb,opt}$ maintains guarantees on long-term convergence; the least tight of the bounds, it only non-trivially improves performance (i.e., constrains selection) when the office-trained policies are deployed in the maze environment.
However, even some assumptions about the structure of unseen space are helpful for more strongly constraining selection.
Though they may violate \emph{theoretical} guarantees on asymptotic performance in general, we show that using the weighted lower bound cost $C_{\text{\tiny{}p=0.5}}^\text{\tiny{}lb,wgt}$, with a $p=50\%$ likelihood a shorter alternative path exists, and the non-simply-connected assumptions result in improvements over the unconstrained UCB bandit in both office environments.

\section{Conclusion and Future Work} \label{sec:conclusion}
We present a data-efficient policy selection approach that leverages Learning over Subgoals Planning-enabled offline alt-policy replay to compute a lower bound on the performance of policies based on the partial map and images collected from the environment during navigation, and use bandit-like method to identify the best-performing policy quickly.
Our approach enables the learning-guided robot to reduce average navigation cost in a wide variety of partially-mapped environments by picking only those policies that are known to perform better or have the potential to do so and thereby significantly reducing the cumulative regret compared to the baseline UCB bandit.
In future, we hope to extend our work to perform policy selection with online retraining or adaptation of policies in new environments.

\section{Acknowledgement}
We would like to thank Jana Košecká, George Konidaris and Kevin Doherty for their thoughtful feedback on this work.
This material is based upon work supported by the National Science Foundation under Grant No.\ 2232733.

\bibliography{references.bib}

\begin{thebibliography}{10}
\providecommand{\url}[1]{#1}
\csname url@samestyle\endcsname
\providecommand{\newblock}{\relax}
\providecommand{\bibinfo}[2]{#2}
\providecommand{\BIBentrySTDinterwordspacing}{\spaceskip=0pt\relax}
\providecommand{\BIBentryALTinterwordstretchfactor}{4}
\providecommand{\BIBentryALTinterwordspacing}{\spaceskip=\fontdimen2\font plus
\BIBentryALTinterwordstretchfactor\fontdimen3\font minus
  \fontdimen4\font\relax}
\providecommand{\BIBforeignlanguage}[2]{{%
\expandafter\ifx\csname l@#1\endcsname\relax
\typeout{** WARNING: IEEEtran.bst: No hyphenation pattern has been}%
\typeout{** loaded for the language `#1'. Using the pattern for}%
\typeout{** the default language instead.}%
\else
\language=\csname l@#1\endcsname
\fi
#2}}
\providecommand{\BIBdecl}{\relax}
\BIBdecl

\bibitem{kaelbling1998planning}
L.~P. Kaelbling, M.~L. Littman, and A.~R. Cassandra, ``Planning and acting in
  partially observable stochastic domains,'' \emph{Artificial Intelligence},
  1998.

\bibitem{stein2018learning}
G.~J. Stein, C.~Bradley, and N.~Roy, ``Learning over subgoals for efficient
  navigation of structured, unknown environments,'' in \emph{Conference on
  Robot Learning}.\hskip 1em plus 0.5em minus 0.4em\relax PMLR, 2018.

\bibitem{richter2014high}
C.~Richter, J.~Ware, and N.~Roy, ``High-speed autonomous navigation of unknown
  environments using learned probabilities of collision,'' in \emph{2014 IEEE
  International Conference on Robotics and Automation (ICRA)}, 2014.

\bibitem{wayne2018unsupervised_merlin}
G.~Wayne, C.-C. Hung, D.~Amos, M.~Mirza, A.~Ahuja, A.~Grabska-Barwinska,
  J.~Rae, P.~Mirowski, J.~Z. Leibo, A.~Santoro \emph{et~al.}, ``Unsupervised
  predictive memory in a goal-directed agent,'' \emph{arXiv preprint
  arXiv:1803.10760}, 2018.

\bibitem{zhu2017target}
Y.~Zhu, R.~Mottaghi, E.~Kolve, J.~J. Lim, A.~Gupta, L.~Fei-Fei, and A.~Farhadi,
  ``Target-driven visual navigation in indoor scenes using deep reinforcement
  learning,'' in \emph{2017 IEEE International Conference on Robotics and
  Automation (ICRA)}, 2017.

\bibitem{kahn2018self}
G.~Kahn, A.~Villaflor, B.~Ding, P.~Abbeel, and S.~Levine, ``Self-supervised
  deep reinforcement learning with generalized computation graphs for robot
  navigation,'' in \emph{2018 IEEE International Conference on Robotics and
  Automation (ICRA)}, 2018.

\bibitem{mirowski2016learning}
P.~Mirowski, R.~Pascanu, F.~Viola, H.~Soyer, A.~J. Ballard, A.~Banino,
  M.~Denil, R.~Goroshin, L.~Sifre \emph{et~al.}, ``Learning to navigate in
  complex environments,'' \emph{arXiv preprint arXiv:1611.03673}, 2016.

\bibitem{lee2021online}
J.~Lee, A.~Pacchiano, V.~Muthukumar, W.~Kong, and E.~Brunskill, ``Online model
  selection for reinforcement learning with function approximation,'' in
  \emph{International Conference on Artificial Intelligence and
  Statistics}.\hskip 1em plus 0.5em minus 0.4em\relax PMLR, 2021.

\bibitem{reisinger2008online}
J.~Reisinger, P.~Stone, and R.~Miikkulainen, ``Online kernel selection for
  bayesian reinforcement learning,'' in \emph{Proceedings of the 25th
  International Conference on Machine Learning}, 2008.

\bibitem{lattimore2020bandit}
T.~Lattimore and C.~Szepesv{\'a}ri, \emph{Bandit Algorithms}.\hskip 1em plus
  0.5em minus 0.4em\relax Cambridge University Press, 2020.

\bibitem{ucb_lai1985asymptotically}
T.~L. Lai, H.~Robbins \emph{et~al.}, ``Asymptotically efficient adaptive
  allocation rules,'' \emph{Advances in Applied Mathematics}, 1985.

\bibitem{gittins1979bandit}
J.~C. Gittins, ``Bandit processes and dynamic allocation indices,''
  \emph{Journal of the Royal Statistical Society: Series B (Methodological)},
  1979.

\bibitem{thompson1933likelihood}
W.~R. Thompson, ``On the likelihood that one unknown probability exceeds
  another in view of the evidence of two samples,'' \emph{Biometrika}, 1933.

\bibitem{kuleshov2014algorithms}
V.~Kuleshov and D.~Precup, ``Algorithms for multi-armed bandit problems,''
  \emph{arXiv preprint arXiv:1402.6028}, 2014.

\bibitem{mannor2004sample}
S.~Mannor and J.~N. Tsitsiklis, ``The sample complexity of exploration in the
  multi-armed bandit problem,'' \emph{Journal of Machine Learning Research},
  2004.

\bibitem{mallozzi2019runtime}
P.~Mallozzi, E.~Castellano, P.~Pelliccione, G.~Schneider, and K.~Tei, ``A
  runtime monitoring framework to enforce invariants on reinforcement learning
  agents exploring complex environments,'' in \emph{2019 IEEE/ACM 2nd
  International Workshop on Robotics Software Engineering (RoSE)}, 2019.

\bibitem{kulhanek2019vision}
J.~Kulh{\'a}nek, E.~Derner, T.~De~Bruin, and R.~Babu{\v{s}}ka, ``Vision-based
  navigation using deep reinforcement learning,'' in \emph{2019 European
  Conference on Mobile Robots (ECMR)}, 2019.

\bibitem{henderson2018rlmatters}
P.~Henderson, R.~Islam, P.~Bachman, J.~Pineau, D.~Precup, and D.~Meger, ``Deep
  reinforcement learning that matters,'' in \emph{AAAI Conference on Artificial
  Intelligence}, 2018.

\bibitem{huang2017adversarial}
S.~Huang, N.~Papernot, I.~Goodfellow, Y.~Duan, and P.~Abbeel, ``Adversarial
  attacks on neural network policies,'' \emph{arXiv preprint arXiv:1702.02284},
  2017.

\bibitem{xiang2018pca}
Y.~Xiang, W.~Niu, J.~Liu, T.~Chen, and Z.~Han, ``A {PCA-based} model to predict
  adversarial examples on {Q-learning} of path finding,'' in \emph{2018 IEEE
  Third International Conference on Data Science in Cyberspace (DSC)}, 2018.

\bibitem{littman1997}
M.~L. Littman, A.~R. Cassandra, and L.~P. Kaelbling, \emph{Learning Policies
  for Partially Observable Environments: Scaling Up}.\hskip 1em plus 0.5em
  minus 0.4em\relax Morgan Kaufmann Publishers Inc., 1997.

\bibitem{candido2011minimum}
S.~Candido and S.~Hutchinson, ``Minimum uncertainty robot navigation using
  information-guided {POMDP} planning,'' in \emph{2011 IEEE International
  Conference on Robotics and Automation}, 2011.

\bibitem{kurniawati2011motion}
H.~Kurniawati, Y.~Du, D.~Hsu, and W.~S. Lee, ``Motion planning under
  uncertainty for robotic tasks with long time horizons,'' \emph{The
  International Journal of Robotics Research}, 2011.

\bibitem{gupta2017cognitive}
S.~Gupta, J.~Davidson, S.~Levine, R.~Sukthankar, and J.~Malik, ``Cognitive
  mapping and planning for visual navigation,'' in \emph{Proceedings of the
  IEEE Conference on Computer Vision and Pattern Recognition (CVPR)}, July
  2017.

\bibitem{zhang2017deep}
J.~Zhang, J.~T. Springenberg, J.~Boedecker, and W.~Burgard, ``Deep
  reinforcement learning with successor features for navigation across similar
  environments,'' in \emph{2017 IEEE/RSJ International Conference on
  Intelligent Robots and Systems (IROS)}, 2017.

\bibitem{tai2017virtualtoreal}
L.~Tai, G.~Paolo, and M.~Liu, ``Virtual-to-real deep reinforcement learning:
  Continuous control of mobile robots for mapless navigation,'' in \emph{2017
  IEEE/RSJ International Conference on Intelligent Robots and Systems (IROS)},
  2017.

\bibitem{ghosh2022model}
A.~Ghosh and S.~R. Chowdhury, ``Model selection in reinforcement learning with
  general function approximations,'' \emph{arXiv preprint arXiv:2207.02992},
  2022.

\bibitem{pacchiano2020regret}
A.~Pacchiano, C.~Dann, C.~Gentile, and P.~Bartlett, ``Regret bound balancing
  and elimination for model selection in bandits and {RL},'' \emph{arXiv
  preprint arXiv:2012.13045}, 2020.

\bibitem{Rahman_2021}
Q.~M. Rahman, P.~Corke, and F.~Dayoub, ``Run-time monitoring of machine
  learning for robotic perception: A survey of emerging trends,'' \emph{IEEE
  Access}, 2021.

\bibitem{zhou2019automated}
W.~Zhou, J.~S. Berrio, S.~Worrall, and E.~Nebot, ``Automated evaluation of
  semantic segmentation robustness for autonomous driving,'' \emph{IEEE
  Transactions on Intelligent Transportation Systems}, 2019.

\bibitem{daftry2016introspective}
S.~Daftry, S.~Zeng, J.~A. Bagnell, and M.~Hebert, ``Introspective perception:
  Learning to predict failures in vision systems,'' in \emph{2016 IEEE/RSJ
  International Conference on Intelligent Robots and Systems (IROS)}, 2016.

\bibitem{merlin2020locally_lomdp}
M.~Merlin, N.~Parikh, E.~Rosen, and G.~Konidaris, ``Locally observable markov
  decision processes,'' in \emph{ICRA 2020 Workshop on Perception, Action,
  Learning}, 2020.

\bibitem{sutton2018reinforcement}
R.~S. Sutton and A.~G. Barto, \emph{Reinforcement Learning: An
  Introduction}.\hskip 1em plus 0.5em minus 0.4em\relax MIT Press, 2018.

\bibitem{hart1968formal_astar}
P.~E. Hart, N.~J. Nilsson, and B.~Raphael, ``A formal basis for the heuristic
  determination of minimum cost paths,'' \emph{IEEE Transactions on Systems
  Science and Cybernetics}, 1968.

\bibitem{bradley2021learning}
C.~Bradley, A.~Pacheck, G.~J. Stein, S.~Castro, H.~Kress-Gazit, and N.~Roy,
  ``Learning and planning for temporally extended tasks in unknown
  environments,'' in \emph{2021 IEEE International Conference on Robotics and
  Automation (ICRA)}, 2021.

\end{thebibliography}

\end{document}